\documentclass{article}
\usepackage{ijcai17}

\usepackage{times}
\usepackage{helvet}
\usepackage{courier}
\usepackage{amssymb}
\usepackage{amsmath}
\usepackage{multirow}
\usepackage{amsopn}
\usepackage{epsfig}
\usepackage{graphicx,subfigure}
\usepackage{array}
\usepackage{url}
\usepackage{bm}
\usepackage{tabularx}
\usepackage{caption}
\usepackage{latexsym}
\usepackage{epstopdf}
\usepackage[export]{adjustbox}
\usepackage{subfig}
\usepackage{tikz}
\usetikzlibrary{fit,positioning}

\title{Semi-supervised Bayesian Deep Multi-modal Emotion Recognition}

\author{Changde Du$^{1}$, Changying Du$^{2}$, Jinpeng Li$^{1}$, Wei-long Zheng$^3$, Bao-liang Lu $^3$, Huiguang He$^{1}$\\
$^1$Research Center for Brain-Inspired Intelligence,\\ Institute of Automation, Chinese Academy of Sciences (CAS), Beijing, China\\
$^2$Laboratory of Parallel Software and Computational Science, Institute of Software, CAS, Beijing, China\\
$^3$Department of Computer Science and Engineering, Shanghai Jiao Tong University, Shanghai, China\\
\{duchangde2016, huiguang.he\}@ia.ac.cn
}

\begin{document}

\maketitle

\begin{abstract}

In emotion recognition, it is difficult to recognize human's emotional states using just a single modality.  Besides, the annotation of physiological emotional data is particularly expensive. These two aspects make the building of effective emotion recognition model challenging. In this paper, we first build a multi-view deep generative model to simulate the generative process of multi-modality emotional data.  By imposing a mixture of Gaussians assumption on the posterior approximation of the latent variables, our model can learn the shared deep representation from multiple modalities. To solve the labeled-data-scarcity problem, we further extend our multi-view model to semi-supervised learning scenario by casting the semi-supervised classification problem as a specialized missing data imputation task. Our semi-supervised multi-view deep generative framework can leverage both labeled and unlabeled data from multiple modalities, where the weight factor for each modality can be learned automatically. Compared with previous emotion recognition methods, our method is more robust and flexible. The experiments conducted on two real multi-modal emotion datasets have demonstrated the superiority of our framework over a number of competitors.
\end{abstract}

\section{Introduction}
With the development of human-computer interaction, emotion recognition has become increasingly important. Since human's emotion contains many nonverbal cues, various modalities ranging from facial expressions, voice, Electroencephalogram (EEG), eye movements to other physiological signals can be used as the indicators of emotional states~\cite{calvo2010affect}. In real-world applications, it is difficult to recognize the emotional states using just a single modality, because signals from different modalities represent different aspects of emotion and provide complementary information.  Recent studies show that integrating multiple modalities can significantly boost the emotion recognition accuracy~\cite{verma2014multimodal,pang2015deep,lu2015combining,liu2016multimodal,soleymani2016analysis,zhang2016enhanced}.

The most successful approach to fuse the information from multiple modalities is based on deep multi-view representation learning~\cite{ngiam2011multimodal,andrew2013deep,srivastava2014multimodal,wang2015deep,chandar2016correlational}. For example, ~\cite{pang2015deep} proposed to learn a joint density model for emotion analysis with a multi-modal Deep Boltzmann Machine (DBM)~\cite{srivastava2014multimodal}. This multi-modal DBM is exploited to model the joint distribution over visual, auditory, and textual features. ~\cite{liu2016multimodal} proposed a multi-modal emotion recognition method by using multi-modal Deep Autoencoders (DAE)~\cite{ngiam2011multimodal}, in which the joint representations of EEG and eye movement signals were extracted.  Nevertheless, there are still limitations with these deep multi-modal emotion recognition methods, e.g., their performances depend on the amount of labeled data.

By using the modern sensor equipments, we can easily collect massive physiological signals, which are closely related to people's emotional states. Despite the convenience of data acquisition, the data labeling procedure requires lots of manual efforts.  Therefore, in most cases only a small set of labeled samples is available, while the majority of whole dataset is left unlabeled. Traditional emotion recognition approaches only utilized the limited amount of labeled data, which may result in severe overfitting.  The most attractive way to deal with this issue is based on Semi-supervised Learning (SSL), which builds more robust model by exploiting both labeled and unlabeled data~\cite{schels2014using,jia2014novel,zhang2016enhanced}. Though multi-modal approaches have been widely implemented for emotion recognition, very few of them explored SSL simultaneously. To the best of our knowledge, only ~\cite{zhang2016enhanced} proposed an enhanced multi-modal co-training algorithm for semi-supervised emotion recognition, but its shallow structure is hard to capture the high-level correlation between different modalities.


Amongst existing SSL approaches, the most competitive one is based on deep generative models, which employs the Deep Neural Networks (DNNs) to learn discriminative features and casts the semi-supervised classification problem as a specialized missing data imputation task. ~\cite{kingma2014semi} and ~\cite{maaloe2016auxiliary} have shown that deep generative models and approximate Bayesian inference exploiting recent advances in scalable variational methods~\cite{VAE,rezende2014stochastic} can provide state-of-the-art performance for semi-supervised classification. Though the Variational Autoencoder (VAE) framework~\cite{VAE} has shown great advantages in SSL, its potential merits remain under-explored. For example, until recently, there was no successful multi-view extension for it. The main difficulty lies in its inherent assumption that the posterior approximation should be conditioned on the data point, which is natural to single-view data but becomes problematic for multi-view case.

In this paper, we propose a novel semi-supervised multi-view deep generative framework for multi-modal emotion recognition. Our framework combines the advantages of deep multi-view representation learning and Bayesian modeling, thus it has sufficient flexibility and robustness in learning joint features and classifier. Our main contributions can be summarized as follows.
\begin{itemize}
\item We propose a multi-view extension for VAE by imposing a mixture of Gaussians assumption on the posterior approximation of the latent variables. For multi-view learning, this is critical for fully exploiting the information from multiple views.
\item We introduce a semi-supervised multi-modal emotion recognition framework based on multi-view VAE. Our framework can leverage both labeled and unlabeled samples from multiple modalities and the weight factor for each modality can be learned automatically, which is critical for building a robust emotion recognition system.
\item We demonstrate the superiority of our framework and provide insightful observations on two real multi-modal emotion datasets.
\end{itemize}

\section{ Multi-view Variational Autoencoder for Semi-supervised Emotion Recognition}
The VAE framework has recently been introduced as a robust model for latent feature learning~\cite{VAE,rezende2014stochastic}.
However, the single-view architecture in VAE can't effectively deal with multi-view data.
In this section, we first build a multi-view VAE, which can learn the shared deep representation from multi-view data. And then, we extend it to the semi-supervised scenario. Assume we are faced with multi-view data that appears as pairs $(\mathfrak{X},\  y) = (\{\mathrm{\mathbf{x}}^{(v)}\}_{v=1}^{V},\  y)$, with  observation $\mathrm{\mathbf{x}}^{(v)}$ from the $v$-th view and the corresponding class label $y$.


\subsection{Multi-view Variational Autoencoder}
\subsubsection{DNN-parameterized Likelihoods}
We assume the latent variable $\mathrm{\mathbf{z}}$  can generate multi-view features $\{\mathrm{\mathbf{x}}^{(v)}\}_{v=1}^V$.  Specifically, we assume $\mathrm{\mathbf{z}}$ generates $\mathrm{\mathbf{x}}^{(v)}$ for any $v\in\{1,...,V\}$, with the following generative model (cf. Fig. \ref{fig:MVAE}a):
\begin{align}
 p_{\theta^{(v)}}(\mathrm{\mathbf{x}}^{(v)}|\mathrm{\mathbf{z}})  &= f(\mathrm{\mathbf{x}}^{(v)}; \mathrm{\mathbf{z}}, \theta^{(v)}),
\end{align}
where $f(\mathrm{\mathbf{x}}^{(v)}; \mathrm{\mathbf{z}}, \theta^{(v)})$ is a suitable likelihood function (e.g. a Gaussian for continuous observation or Bernoulli for binary observation), which is formed by a non-linear transformation of the latent variable $\mathbf{z}$. This non-linear transformation is essential to allow for higher moments
of the data to be captured by the density model, and we choose these non-linear functions to be DNNs, referred to as the generative networks, with parameters $\{\theta^{(v)}\}_{v=1}^{V}$. Note that, the likelihoods for different data views are assumed to be independent of each other, with different nonlinear transformations.

The Bayesian Canonical Correlation Analysis (CCA) model ~\cite{Klami2013Bayesian} can be seen as a special case of our model, where linear shallow transformations were used to generate each data view and only two different views were considered. ~\cite{Wang2016Deep} used a similar deep nonlinear generative process as ours to construct deep Bayesian CCA model, but during inference they construct the variational posterior approximation from just one view and ignore the rest one. Such a choice is convenient for inference and computation, but only seeks suboptimal solutions as it doesn't fully exploit the data.  As shown in the following, we assume the variational approximation to the posterior of latent variables to be a mixture of Gaussians, utilizing information from multiple views.

\subsubsection{Gaussian Prior and Mixture of Gaussians Posterior}

Typically, both the prior $p(\mathrm{\mathbf{z}})$ and the approximate posterior  $q_{\phi}(\mathrm{\mathbf{z}}|\mathfrak{X})$ are assumed to be Gaussian distributions~\cite{VAE,rezende2014stochastic} in order to maintain mathematical and computational tractability. Although
this assumption has leaded to favorable results on several tasks, it is clearly a restrictive and often unrealistic assumption. Specifically, the choice of a Gaussian distribution for $p(\mathrm{\mathbf{z}})$ and $q_{\phi}(\mathrm{\mathbf{z}}|\mathfrak{X})$ imposes a strong uni-modal structure assumption on the latent space. However, for data distributions that are strongly multi-modal, the uni-modal Gaussian assumption inhibits the model's ability to extract
and represent important structure in the data.
To improve the flexibility of the model, one way is to impose a mixture of Gaussians assumption on $p(\mathrm{\mathbf{z}})$. However, it has the risk of creating separate ``islands" of discontinuous manifolds that may break the meaningfulness of the representation in the latent space.

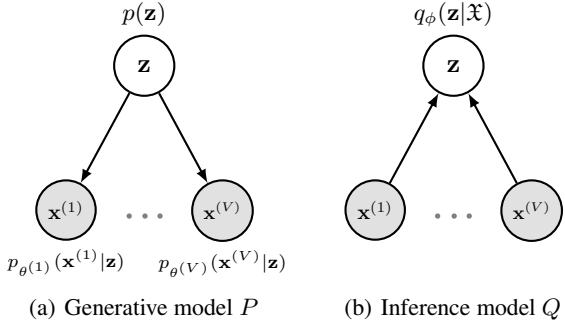
\begin{figure}[t]
  \centering
  \subfigure[Generative model $P$] {
            \begin{tikzpicture}[line width = 0pt]
            \tikzstyle{main}=[circle, thick, draw =black!100, node distance = 11mm]
            \tikzstyle{connect}=[-latex, thick]
            \tikzstyle{box}=[rectangle, draw=black!100]
              \node[main,  minimum size = 8mm] (z) [label=above:\footnotesize  $p(\mathrm{\mathbf{z}})$] {$\mathrm{\mathbf{z}}$ };
              \node[main, scale=0.8, fill = black!12] (x_1) [below=of z,xshift=-1.3cm, label=below:\scriptsize {$p_{\theta^{(1)}}(\mathrm{\mathbf{x}}^{(1)}|\mathrm{\mathbf{z}})$}] {$\mathrm{\mathbf{x}}^{(1)}$};
              \node[main, scale=0.8, fill = black!12] (x_V) [below=of z,xshift=1.3cm, label=below:\scriptsize {$p_{\theta^{(V)}}(\mathrm{\mathbf{x}}^{(V)}| \mathrm{\mathbf{z}})$}] {\small $\mathrm{\mathbf{x}}^{(V)}$ };
              \fill[fill = black!50] (-0.2,-2.0)circle(1pt) (0.01,-2.0)circle(1pt) (0.21,-2.0)circle(1pt);
              \path (z) edge [connect] (x_1)
                    (z) edge [connect] (x_V);
            \coordinate (ne)at(current bounding box.north east);
            \coordinate (sw)at(current bounding box.south west);
            \end{tikzpicture}}
\hspace {-0.1cm}
  \subfigure [Inference model $Q$] {
            \begin{tikzpicture}[line width = 0pt]
            \tikzstyle{main}=[circle,  thick, draw =black!100, node distance = 11mm]
            \tikzstyle{connect}=[-latex, thick]
            \tikzstyle{box}=[rectangle, draw=black!100]
            \useasboundingbox(ne)rectangle(sw);
              \node[main, minimum size = 8mm] (z) [label=above: \footnotesize {$q_\phi(\mathrm{\mathbf{z}}|\mathfrak{X})$}] {$\mathrm{\mathbf{z}}$ };
              \node[main, scale=0.8, fill = black!12] (x_1) [below=of z,xshift=-1.3cm,label=below:\footnotesize] {$\mathrm{\mathbf{x}}^{(1)}$};
              \node[main, scale=0.8, fill = black!12] (x_V) [below=of z,xshift=1.3cm, label=below:\footnotesize] {\small$\mathrm{\mathbf{x}}^{(V)}$ };
              \fill[fill = black!50] (-0.21,-2.0)circle(1pt) (0.01,-2.0)circle(1pt) (0.21,-2.0)circle(1pt);
              \path (x_1) edge [connect] (z)
                    (x_V) edge [connect] (z);
            \end{tikzpicture}}
\vspace {-0.3cm}
\caption{Graphical model of the multi-view VAE, where $\mathfrak{X} = \{\mathrm{\mathbf{x}}^{(v)}\}_{v=1}^V $.}
\label{fig:MVAE}
\end{figure}

To learn more powerful and expressive models -- in particular, models with multi-modal latent variable structures for multi-modal emotion recognition
applications -- we seek a mixture of Gaussians for $q_{\phi}(\mathrm{\mathbf{z}}|\mathfrak{X})$, while preserving $p(\mathrm{\mathbf{z}})$ as a standard Gaussian. Thus (cf. Fig. \ref{fig:MVAE}b),
\begin{align}\label{Posterior}
\nonumber p(\mathrm{\mathbf{z}}) &= \mathcal{N}\left(\mathrm{\mathbf{z}}|\bm{0}, \mathrm{\mathbf{I}}\right),\\
q_\phi(\mathrm{\mathbf{z}}|\mathfrak{X})  & = \sum\limits_{v=1}^{V}\lambda^{(v)}\mathcal{N}\left(\mathrm{\mathbf{z}}|\bm{\mu}_{\phi^{(v)}}(\mathrm{\mathbf{x}}^{(v)}),\ \bm{\Sigma}_{\phi^{(v)}}(\mathrm{\mathbf{x}}^{(v)}) \right),
\end{align}
where the mean $\bm{\mu}_{\phi^{(v)}}$ and the covariance $\bm{\Sigma}_{\phi^{(v)}}$ are nonlinear functions of the observation $\mathrm{\mathbf{x}}^{(v)}$, with variational parameter $\bm{\phi}^{(v)}$. As in our generative model, we choose these nonlinear functions to be DNNs, referred to as the inference networks. $\lambda^{(v)}$ is the non-negative normalized weight factor for the $v$-th view, i.e., $\lambda^{(v)} > 0$ and $\sum_{v=1}^V\lambda^{(v)} =1$.   By conditioning the posterior approximation on the data point, we avoid variational parameters per data point, instead only requiring to fit global variational parameters. Note that, our mixed Gaussian assumption on the variational approximation distinguishes our method from all existing ones using the auto-encoding variational framework ~\cite{VAE,Wang2016Deep,Burda2016Importance,Kingma2016Improving,serban2016multi,maaloe2016auxiliary}. For multi-view learning, this is critical for fully exploiting the information from multiple views.

\subsection{Semi-supervised Emotion Recognition}
In semi-supervised classification, only a subset of the samples have corresponding class labels, and we focus on using the multi-view VAE to build a model (semiMVAE) that learns classifier from both labeled and unlabeled multi-view data.  Since the emotional data is continuous, we choose the Gaussian likelihoods.
Then the generative model $P$ is defined as $p(y)p(\mathrm{\mathbf{z}})\prod_{v=1}^{V}p_{\theta^{(v)}}(\mathrm{\mathbf{x}}^{(v)}|y, \mathrm{\mathbf{z}})$ (cf. Fig. 2a):
\begin{align}
\nonumber p(y) &= \mathrm{Cat}\left(y|\bm{\pi}\right),\\
\nonumber p(\mathrm{\mathbf{z}}) &= \mathcal{N}\left(\mathrm{\mathbf{z}}|\bm{0}, \mathrm{\mathbf{I}}\right),\\
p_{\theta^{(v)}}(\mathrm{\mathbf{x}}^{(v)}|y, \mathrm{\mathbf{z}}) & = \mathcal{N}\left(\bm{\mu}_{\theta^{(v)}}(y, \mathbf{z}),\ \mathrm{diag}(\bm{\sigma}^{2}_{\theta^{(v)}}(y, \mathbf{z})) \right),
\end{align}
where $\mathrm{Cat}(\cdot)$ denotes the categorical distribution, $y$ is treated as a latent variable for the unlabeled data points,
and the mean $\bm{\mu}_{\theta^{(v)}}$ and variance $\bm{\sigma}^{2}_{\theta^{(v)}}$ are nonlinear functions of $y$ and $\mathbf{z}$, with parameter $\theta^{(v)}$.
The inference model $Q$ is defined as $q_{\varphi}(y|\mathfrak{X})q_\phi(\mathrm{\mathbf{z}}|\mathfrak{X}, y)$ (cf. Fig. 2b):
\begin{align}
 q_{\varphi}(y|\mathfrak{X}) &= \mathrm{Cat}\left(y|\bm{\pi}_{\varphi}(\mathfrak{X})\right),\\
\nonumber q_\phi(\mathrm{\mathbf{z}}|\mathfrak{X}, y)  &= \sum\limits_{v=1}^{V}\lambda^{(v)}\mathcal{N}\left(\mathrm{\mathbf{z}}|\bm{\mu}_{\phi^{(v)}}(\mathrm{\mathbf{x}}^{(v)}, y),\ \bm{\Sigma}_{\phi^{(v)}}(\mathrm{\mathbf{x}}^{(v)}, y) \right),
\end{align}
where $q_\phi(\mathrm{\mathbf{z}}|\mathfrak{X}, y)$ is assumed to be a mixture of Gaussians to combine the information from multiple data views. Intuitively,  $q_\phi(\mathrm{\mathbf{z}}|\mathfrak{X}, y)$, $p_{\theta^{(v)}}(\mathrm{\mathbf{x}}^{(v)}|y, \mathrm{\mathbf{z}})$ and $q_{\varphi}(y|\mathfrak{X})$ correspond to the encoder, the decoder and the classifier, respectively.

For brevity, we omit the explicit dependencies on $\mathrm{\mathbf{x}}^{(v)}$, $y$ and $\mathrm{\mathbf{z}}$ for the moment variables mentioned above hereafter. In principle, $\bm{\mu}_{\theta^{(v)}}$, $\bm{\sigma}^{2}_{\theta^{(v)}}$, $\bm{\pi}_{\varphi}$, $\bm{\mu}_{\phi^{(v)}}$ and $\bm{\Sigma}_{\phi^{(v)}}$  can be implemented by various DNN models, e.g., Multiple Layer Perceptrons (MLP) and Convolutional Neural Networks (CNN).

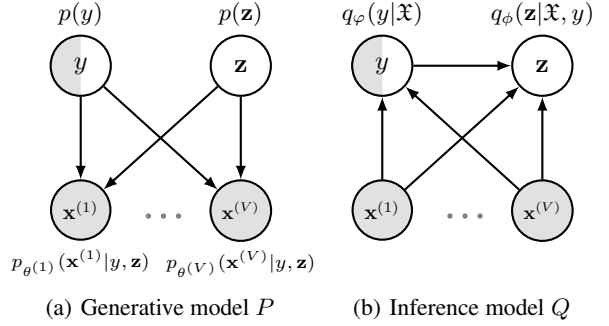
\begin{figure}[t]
  \centering
    \subfigure[Generative model $P$] {
            \begin{tikzpicture}[line width = 0pt]
            \tikzstyle{main}=[circle, thick, draw =black!100, node distance = 11mm]
            \tikzstyle{connect}=[-latex, thick]
            \tikzstyle{box}=[rectangle, draw=black!100]
              \draw[fill = black!12] (-0.0,-0.4) arc (-90:-270:0.4cm);
              \node[main, minimum size = 8mm] (y) [label=above:\footnotesize  $p(y)$] { $y$};
              \node[main, minimum size = 8mm] (z) [right=of y, xshift=0.2cm,label=above:\footnotesize  $p(\mathrm{\mathbf{z}})$] {$\mathrm{\mathbf{z}}$ };
              \node[main,  scale=0.8, fill = black!12] (x_1) [below=of y,label=below:\scriptsize {$p_{\theta^{(1)}}(\mathrm{\mathbf{x}}^{(1)}|y, \mathrm{\mathbf{z}})$}] {$\mathrm{\mathbf{x}}^{(1)}$};
              \node[main,  scale=0.8, fill = black!12] (x_V) [below=of z,label=below:\scriptsize {$p_{\theta^{(V)}}(\mathrm{\mathbf{x}}^{(V)}|y, \mathrm{\mathbf{z}})$}] {\small $\mathrm{\mathbf{x}}^{(V)}$ };
              \fill[fill = black!50] (0.9,-2.0)circle(1pt) (1.1,-2.0)circle(1pt) (1.3,-2.0)circle(1pt);
              \path (y) edge [connect] (x_1)
                    (z) edge [connect] (x_1)
		            (y) edge [connect] (x_V)
                    (z) edge [connect] (x_V);
            \coordinate (ne)at(current bounding box.north east);
            \coordinate (sw)at(current bounding box.south west);
            \end{tikzpicture}}
\hspace {-0.53cm}
  \subfigure [Inference model $Q$] {
            \begin{tikzpicture}[line width = 0pt]
            \tikzstyle{main}=[circle, thick, draw =black!100, node distance = 11mm]
            \tikzstyle{connect}=[-latex, thick]
            \tikzstyle{box}=[rectangle, draw=black!100]
            \useasboundingbox(ne)rectangle(sw);
              \draw[fill = black!12] (-0.0,-0.4) arc (-90:-270:0.4cm);
              \node[main, minimum size = 8mm] (y) [label=above:\footnotesize $q_{\varphi}(y|\mathfrak{X})$] {$y$};
              \node[main, minimum size = 8mm] (z) [right=of y, xshift=0.2cm,label=above: \footnotesize {$q_\phi(\mathrm{\mathbf{z}}|\mathfrak{X}, y)$}] {$\mathrm{\mathbf{z}}$ };
              \node[main,  scale=0.8, fill = black!12] (x_1) [below=of y,label=below:\footnotesize] {$\mathrm{\mathbf{x}}^{(1)}$};
              \node[main,  scale=0.8, fill = black!12] (x_V) [below=of z,label=below:\footnotesize] {\small$\mathrm{\mathbf{x}}^{(V)}$ };
              \fill[fill = black!50] (0.9,-2.0)circle(1pt) (1.1,-2.0)circle(1pt) (1.3,-2.0)circle(1pt);
              \path (x_1) edge [connect] (y)
                    (x_1) edge [connect] (z)
		            (x_V) edge [connect] (y)
                    (x_V) edge [connect] (z)
                    (y)   edge [connect] (z);
            \end{tikzpicture}}
\vspace {-0.3cm}
\caption{Graphical model of the semiMVAE for semi-supervised multi-view learning, where $\mathfrak{X} = \{\mathrm{\mathbf{x}}^{(v)}\}_{v=1}^V $.}
\label{fig:semiMVAE}
\end{figure}

\subsection{Variational Lower Bound}
The variational lower bound on the marginal likelihood for a single labeled data point is
\begin{align}
\nonumber \log p_{\theta}(\mathfrak{X}, y) & = \log \int_{\mathrm{\mathbf{z}}} p_{\theta}(\mathfrak{X}, y, \mathrm{\mathbf{z}})\ d \mathrm{\mathbf{z}} \\
\nonumber & \ge \mathbb{E}_{q_{\phi}(\mathrm{\mathbf{z}}|\mathfrak{X}, y)}\bigg[\log \frac{ p_{\theta}(\mathfrak{X}, y, \mathrm{\mathbf{z}})} {q_{\phi}(\mathrm{\mathbf{z}}|\mathfrak{X}, y)}\bigg]  \\
\nonumber & = \mathbb{E}_{q_{\phi}(\mathrm{\mathbf{z}}|\mathfrak{X},y)}\bigg[\sum_{v=1}^{V} \log p_{\theta^{(v)}}(\mathrm{\mathbf{x}}^{(v)}|y, \mathrm{\mathbf{z}}) + \log p(y)\\
\nonumber &\hspace{0.4cm} + \log p(\mathrm{\mathbf{z}}) - \log q_{\phi}(\mathrm{\mathbf{z}}|\mathfrak{X},y) \bigg]\\
\nonumber &  \ge \mathbb{E}_{q_{\phi}(\mathrm{\mathbf{z}}|\mathfrak{X},y)}\bigg[\sum_{v=1}^{V} \log p_{\theta^{(v)}}(\mathrm{\mathbf{x}}^{(v)}|y, \mathrm{\mathbf{z}}) + \log p(y)\\
\nonumber &\hspace{0.4cm} + \log p(\mathrm{\mathbf{z}}) \bigg] - \sum_{v=1}^V \lambda^{(v)} \cdot \log \bigg(\sum_{l=1}^V \lambda^{(l)} \cdot \omega_{v, l}\bigg) \\
 & \equiv -\mathcal{L}(\mathfrak{X}, y),
\end{align}
where $\omega_{v, l} = \mathcal{N}\big( \bm{\mu}_{\phi^{(v)}} | \bm{\mu}_{\phi^{(l)}},\ \bm{\Sigma}_{\phi^{(v)}} + \bm{\Sigma}_{\phi^{(l)}} \big)$. Note that, the Shannon entropy $\mathbb{E}_{q_{\phi}(\mathrm{\mathbf{z}}|\mathfrak{X},y)}[- \log q_{\phi}(\mathrm{\mathbf{z}}|\mathfrak{X},y)]$ is hard to compute analytically, and we have used the Jensen's inequality to derive a lower bound of it:
\begin{align}
\nonumber & \mathbb{E}_{q_{\phi}(\mathrm{\mathbf{z}}|\mathfrak{X},y)}[- \log q_{\phi}(\mathrm{\mathbf{z}}|\mathfrak{X},y)] \\
\nonumber  & = - \sum_{v=1}^V \lambda^{(v)} \int \mathcal{N}\big(\mathrm{\mathbf{z}} | \bm{\mu}_{\phi^{(v)}},\ \bm{\Sigma}_{\phi^{(v)}} \big)  \log q_{\phi}(\mathrm{\mathbf{z}}|\mathfrak{X},y)\ d \mathrm{\mathbf{z}} \\
\nonumber  & \geq - \sum_{v=1}^V \lambda^{(v)}  \log \bigg( \sum_{l=1}^V \lambda^{(l)}  \int \mathcal{N}\big(\mathrm{\mathbf{z}} | \bm{\mu}_{\phi^{(v)}},\ \bm{\Sigma}_{\phi^{(v)}} \big) \\
\nonumber &\hspace{0.4cm} \cdot \mathcal{N}\big(\mathrm{\mathbf{z}} | \bm{\mu}_{\phi^{(l)}},\ \bm{\Sigma}_{\phi^{(l)}} \big) \bigg) \ d \mathrm{\mathbf{z}} \\
\nonumber  & = - \sum_{v=1}^V \lambda^{(v)} \log \bigg( \sum_{l=1}^V \lambda^{(l)} \mathcal{N}\big( \bm{\mu}_{\phi^{(v)}} | \bm{\mu}_{\phi^{(l)}}, \bm{\Sigma}_{\phi^{(v)}} + \bm{\Sigma}_{\phi^{(l)}} \big) \bigg).
\end{align}

For unlabeled data, we further introduce the variational distribution $q_{\varphi}(y|\mathfrak{X})$ for $y$:
\begin{align}
\nonumber \log p_{\theta}(\mathfrak{X}) & = \log \int_{\mathrm{\mathbf{z}}} \int_{y} p_{\theta}(\mathfrak{X}, y, \mathrm{\mathbf{z}})\ d y\ d \mathrm{\mathbf{z}} \\
\nonumber & \ge \mathbb{E}_{q_{\varphi, \phi}(y, \mathrm{\mathbf{z}}|\mathfrak{X})}\bigg[\log \frac{ p_{\theta}(\mathfrak{X}, y, \mathrm{\mathbf{z}})} {q_{\varphi, \phi}(y, \mathrm{\mathbf{z}}|\mathfrak{X})}\bigg]  \\
\nonumber & = \mathbb{E}_{q_{\varphi}(y|\mathfrak{X})}\big[-\mathcal{L}(\mathfrak{X},y)- \log q_{\varphi}(y|\mathfrak{X})\big] \\
 & \equiv -\mathcal{U}(\mathfrak{X}),
\end{align}
with $q_{\varphi, \phi}(y, \mathrm{\mathbf{z}}|\mathfrak{X}) = q_{\varphi}(y|\mathfrak{X}) q_{\phi}(\mathrm{\mathbf{z}}|\mathfrak{X}, y)$.
The objective function for the entire dataset is now:
\begin{equation}
\begin{split}
 \mathcal{J} & = \sum_{(\mathfrak{X},y) \in S_{l}}\mathcal{L}(\mathfrak{X},y) + \sum_{\mathfrak{X} \in S_{u}}\mathcal{U}(\mathfrak{X}) \,,
\end{split}
\end{equation}
where $S_l$ and $S_u$ are labeled and unlabeled dataset, respectively. The classification accuracy can be improved by introducing
an explicit classification loss for labeled data. The extended objective function is:
\begin{equation}\label{obj}
\begin{split}
\mathcal{F} & = \mathcal{J} + \alpha \cdot \sum_{(\mathfrak{X},y) \in S_{l}}[-\log q_{\varphi}(y|\mathfrak{X})] \,,
\end{split}
\end{equation}
where the  hyper-parameter $\alpha$ is a weight between generative and discriminative learning. We set $\alpha = \beta \cdot (N_l + N_u)$, where $\beta$ is a scaling constant, and $N_l$ and $N_u$ are the numbers of labeled and unlabeled data points in one minibatch, respectively. Note that, the classifier $q_{\varphi}(y|\mathfrak{X})$ is also used at test phase for the prediction of unseen data.

\subsection{Optimization}
Eq. (\ref{obj}) provides a unified objective function for optimizing the parameters of encoder, decoder and classifier networks. This optimization can be done jointly, without resort to the variational EM algorithm, using the stochastic backpropagation technique~\cite{VAE,rezende2014stochastic}.

\subsubsection{Reparameterization Trick}
The reparameterization trick is a vital component of the algorithm, because it allows us to easily take the derivative of  $\mathbb{E}_{q_{\phi}(\mathrm{\mathbf{z}}|\mathfrak{X},y)}[\log p_{\theta^{(v)}}(\mathrm{\mathbf{x}}^{(v)}|y, \mathrm{\mathbf{z}})]$ with respect to the variational parameters $\phi$.
However, the use of a mixture of Gaussians for the variational distribution $q_{\phi}(\mathrm{\mathbf{z}}|\mathfrak{X},y)$ makes the application of reparameterization trick challenging.
It can be shown that, for any $v \in\{1,...,V\}$, $\mathbb{E}_{q_{\phi}(\mathrm{\mathbf{z}}|\mathfrak{X},y)}[\log p_{\theta^{(v)}}(\mathrm{\mathbf{x}}^{(v)}|y, \mathrm{\mathbf{z}})]$  can be rewritten, using the location-scale transformation for the Gaussian distribution, as:
\begin{align}
\label{reparameterization}
\nonumber  & \mathbb{E}_{q_{\phi}(\mathrm{\mathbf{z}}|\mathfrak{X},y)}[\log p_{\theta^{(v)}}(\mathrm{\mathbf{x}}^{(v)}|y, \mathrm{\mathbf{z}})] \\
 &  = \sum_{l=1}^V \lambda^{(l)} \mathbb{E}_{\mathcal{N}(\bm{\epsilon}^{(l)}|\mathrm{\mathbf{0}},\mathrm{\mathbf{I}})}\big[\log p_{\theta^{(v)}} (\mathrm{\mathbf{x}}^{(v)}|y, \bm{\mu}_{\phi^{(l)}} + \mathrm{\mathbf{R}}_{\phi^{(l)}} \bm{\epsilon}^{(l)} ) \big],
\end{align}
where $\mathrm{\mathbf{R}}_{\phi^{(l)}}\mathrm{\mathbf{R}}_{\phi^{(l)}}^\top = \bm{\Sigma}_{\phi^{(l)}}$ and  $l\in\{1,...,V\}$.
\subsubsection{Gradients of the Objective}
While the expectations on the right hand side of Eq. (\ref{reparameterization}) still cannot be solved analytically, their gradients w.r.t. $\theta^{(v)}$, $\phi^{(l)}$ and $\lambda^{(l)}$ can be efficiently estimated using the following Monte-Carlo estimators,
\vskip -0.1in
\begin{align}
\nonumber  &   \frac{\partial}{\partial \theta^{(v)}} \mathbb{E}_{q_{\phi}(\mathrm{\mathbf{z}}|\mathfrak{X},y)}[\log p_{\theta^{(v)}}(\mathrm{\mathbf{x}}^{(v)}|y, \mathrm{\mathbf{z}})] \\
\nonumber  &  \qquad = \sum_{l=1}^V  \lambda^{(l)} \mathbb{E}_{\mathcal{N}(\bm{\epsilon}^{(l)} |\mathrm{\mathbf{0}},\mathrm{\mathbf{I}})}\bigg[\frac{\partial }{\partial \theta^{(v)}} \log p_{\theta^{(v)}}\big (\mathrm{\mathbf{x}}^{(v)}|y, \mathrm{\mathbf{z}}^{(l)} \big )\bigg] \\
  &  \qquad \approx \frac{\lambda^{(l)}}{T}  \sum_{t=1}^T \sum_{l=1}^V \frac{\partial}{\partial \theta^{(v)}} \log p_{\theta^{(v)}} \big(\mathrm{\mathbf{x}}^{(v)}|y, \mathrm{\mathbf{z}}^{(l,t)} \big),
\end{align}
\vskip -0.1in
\begin{align}
\nonumber  &   \frac{\partial}{\partial \phi^{(l)}} \mathbb{E}_{q_{\phi}(\mathrm{\mathbf{z}}|\mathfrak{X},y)}[\log p_{\theta^{(v)}}(\mathrm{\mathbf{x}}^{(v)}|y, \mathrm{\mathbf{z}})] \\
\nonumber  & \qquad  = \lambda^{(l)} \frac{\partial}{\partial \phi^{(l)}} \mathbb{E}_{\mathcal{N}(\bm{\epsilon}^{(l)}|\mathrm{\mathbf{0}},\mathrm{\mathbf{I}})}\Big[\frac{\partial }{\partial \mathrm{\mathbf{z}}^{(l)} } \log p_{\theta^{(v)}}(\mathrm{\mathbf{x}}^{(v)}|y, \mathrm{\mathbf{z}}^{(l)}) \\
\nonumber  & \qquad \hspace{0.4cm} \cdot \Big( \frac{\partial \bm{\mu}_{\phi^{(l)}} }{\partial \phi^{(l)}} + \frac{\partial \mathrm{\mathbf{R}}_{\phi^{(l)}}}{\partial \phi^{(l)}} \bm{\epsilon}^{(l)} \Big) \Big] \\
\nonumber & \qquad  \approx \ \frac{\lambda^{(l)}}{T}\sum_{t=1}^T  \frac{\partial }{\partial \mathrm{\mathbf{z}}^{(l,t)} } \log p_{\theta^{(v)}}(\mathrm{\mathbf{x}}^{(v)}|y, \mathrm{\mathbf{z}}^{(l,t)})      \\
  & \qquad \hspace{0.4cm} \cdot \Big( \frac{\partial \bm{\mu}_{\phi^{(l)}} }{\partial \bm{\phi}^{(l)}} + \frac{\partial \mathrm{\mathbf{R}}_{\phi^{(l)}}}{\partial \phi^{(l)}} \bm{\epsilon}^{(l,t)} \Big), \qquad\qquad
\end{align}
\vskip -0.1in
\begin{align}
\nonumber  &   \frac{\partial}{\partial \lambda^{(l)}} \mathbb{E}_{q_{\phi}(\mathrm{\mathbf{z}}|\mathfrak{X},y)}[\log p_{\theta^{(v)}}(\mathrm{\mathbf{x}}^{(v)}|y, \mathrm{\mathbf{z}})] \\
\nonumber  & \qquad  =  \mathbb{E}_{\mathcal{N}(\bm{\epsilon}^{(l)}|\mathrm{\mathbf{0}},\mathrm{\mathbf{I}})}\big[\log p_{\theta^{(v)}} (\mathrm{\mathbf{x}}^{(v)}|y, \mathrm{\mathbf{z}}^{(l)} ) \big]\\
  &  \qquad \approx  \frac{1}{T}\sum_{t=1}^T  \log p_{\theta^{(v)}} \big(\mathrm{\mathbf{x}}^{(v)}|y, \mathrm{\mathbf{z}}^{(l,t)} \big),  \qquad\qquad\qquad
\end{align}
where $\mathrm{\mathbf{z}}^{(l)}$ is evaluated at $\mathrm{\mathbf{z}}^{(l)} = \bm{\mu}_{\phi^{(l)}} + \mathrm{\mathbf{R}}_{\phi^{(l)}} \bm{\epsilon}^{(l)}$ and $\mathrm{\mathbf{z}}^{(l,t)} = \bm{\mu}_{\phi^{(l)}} + \mathrm{\mathbf{R}}_{\phi^{(l)}} \bm{\epsilon}^{(l,t)}$ with $\bm{\epsilon}^{(l,t)} \sim \mathcal{N}(\mathrm{\mathbf{0}},\mathrm{\mathbf{I}})$. In practice, it suffices to use a small $T$ (e.g. $T = 1$) and then estimate the gradient using minibatches of data points. We use the same random numbers $\bm{\epsilon}^{(l,t)}$ for all estimators to have lower variance. The gradient w.r.t. $\varphi$ is omitted here, since it can be derived straightforwardly by using traditional reparameterization trick~\cite{kingma2014semi}.

The gradients of the loss for semiMVAE (Eq. (\ref{obj})) can then be computed by a direct application of the chain rule and estimators presented above.
During optimization we can use the estimated gradients in conjunction with standard stochastic gradient based
optimization methods such as SGD, RMSprop or Adam~\cite{kingma2014adam}.  Overall, the model can be trained with reparameterization trick for backpropagation through the mixed Gaussian latent variables.

\section{Experiments}
In this section, we present extensive experimental results to demonstrate the effectiveness of the proposed semi-supervised multi-view framework for emotion recognition.
\subsection{Experimental Testbed and Setup}
\noindent \textbf{Data description}\ \ Two multi-modal emotion datasets, the SEED dataset\footnote{http://bcmi.sjtu.edu.cn/\%7Eseed/index.html}~\cite{lu2015combining} and the DEAP dataset\footnote{http://www.eecs.qmul.ac.uk/mmv/datasets/deap/download.html}~\cite{koelstra2012deap}, were used in our experiments.

The SEED dataset contains EEG and eye movement signals from 15 subjects during watching 15 movie clips, where each movie clip lasts about 4 minutes long.
The EEG signals were recorded from 62 channels and the eye movement signals contained information about blink, saccade fixation and so on.  In our experiment, we used the data from 9 subjects across 3 sessions, totally 27 data files. For each data file, data from watching the 1-9 movie clips were used as training set, while data from watching the 10-12 movie clips were used as validation set and the rest (13-15) were used as testing set.

The DEAP dataset contains EEG and peripheral physiological signals of 32 participants. Signals were recorded when they were watching 40 one-minute duration
music videos. The EEG signals were recorded from 32 channels, whereas the peripheral physiological signals were recorded from 8 channels. The participants, using values from 1 to 9, rated each music video in terms of the levels of valence, arousal and so on. In our experiment, the valence-arousal space was divided into four quadrants according to the ratings. The threshold we used was 5, leading to four classes of data. Considering the fuzzy boundary of emotions and the
variations of participants' ratings possibly associated with individual difference in rating scale, we discarded the samples whose ratings of arousal and valence are between 3 and 6. The dataset was randomly divided into 10-folds, where 8 folds for training, one fold for validation and the last fold for testing. The size of testing set is relative small, because some graph-based semi-supervised baselines are hard to deal with large dataset.

\vskip 0.02in
\noindent \textbf{Feature selection}\ \ For SEED dataset, ~\citeauthor{lu2015combining} have extracted the Differential Entropy (DE) features and 33 eye movement features from EEG and eye movement signals~\cite{lu2015combining}.  We also used these features in our experiments.  For DEAP dataset, we extracted the DE features from EEG and peripheral physiological signals. The DE features can be calculated in four frequency bands: theta (4-8Hz), alpha (8-14Hz), beta (14-31Hz), and gamma (31-45Hz), and we used all band's features. The details of the data used in our experiments were summarized in Table \ref{benchmark data sets}.
\begin{table}[!htbp]
\scriptsize
\centering
\setlength\tabcolsep{1.5pt}
\setlength{\abovecaptionskip}{3pt}
\setlength{\belowcaptionskip}{-4pt}
\caption{The details of the datasets used in our experiments.}
\label{benchmark data sets}
\begin{tabular}{|c|c|c|c|c|c|c|}
\hline
 Datasets        & \#Instances           &\#Features                  &\#Training      & \#Validation & \#Testing   & \#Classes     \\ \hline\hline
 SEED            & 22734                 &310(EEG), 33(Eye)           & 13473          & 4725         & 4536        & 3       \\
 DEAP            & 21042                 &128(EEG), 32(Phy.)          & 16834          & 2104         & 2104        & 4       \\ \hline
\end{tabular}
\end{table}
\vskip -0.1in
\noindent \textbf{Compared methods}\ \ We compared our semiMVAE with a broad range of solutions, including supervised learning, transductive and inductive semi-supervised learning.
 We briefly summarize the various baselines in the following.
\begin{itemize}
\item $\mathbf{MAE}$: the multi-view extension of deep autoencoders, which can be used to extract the joint representations from multiple modalities~\cite{ngiam2011multimodal}.
\vspace{-0.04cm}
\item $\mathbf{DCCA}$: the full deep neural network extension of Canonical Correlation Analysis (CCA). DCCA can learn deep nonlinear mappings of two views, which are maximally correlated~\cite{andrew2013deep}.
\vspace{-0.04cm}
\item $\mathbf{DCCAE}$: a deep multi-view representation learning model which combines the advantages of the DCCA and deep autoencoders. In particular, DCCAE consists of two autoencoders and optimizes the combination of canonical correlation between the learned bottleneck representations and the reconstruction errors of the autoencoders ~\cite{wang2015deep}.
\vspace{-0.04cm}
\item $\mathbf{AMMSS}$: a graph-based multi-view semi-supervised classification algorithm, which can integrate heterogeneous features from both labeled and unlabeled data~\cite{Cai2013Heterogeneous}.
\vspace{-0.04cm}
\item $\mathbf{AMGL}$: a latest auto-weighted multiple graph learning framework, which can be applied to multi-view semi-supervised classification task~\cite{nie2016parameter}.
\vspace{-0.04cm}
\item $\mathbf{semiVAE}$: a single-view semi-supervised deep generative model proposed in~\cite{kingma2014semi}. We evaluate semiVAE's performance for each modality and the concatenation of all modalities, respectively.
\end{itemize}
For MAE, DCCA and DCCAE, we used the Support Vector Machines\footnote{http://www.csie.ntu.edu.tw/\%7Ecjlin/liblinear/.} (SVM) and transductive SVM\footnote{http://svmlight.joachims.org/.} (TSVM) for  supervised learning and transductive semi-supervised learning, respectively.
\vskip 0.02in
\noindent \textbf{Parameter setting}\ \ For semiMVAE, we considered multiple layer perceptrons as the type of inference and generative networks. On both datasets, we set the structures of the inference and generative networks for each view as `100-50-30' and `30-50-100', respectively.  We used the Adam optimizer~\cite{kingma2014adam} with a learning rate $\eta = 3\times10^{-4}$ in training. The scaling constant $\beta$ was selected from \{0.1, 0.5, 1\} throughout the experiments. The weight factor for each view  was initialized with $\lambda^{(v)}=1/V$, where $V$ is the number of views. For MAE, DCCA and DCCAE, we considered the same setups (network structure, learning rate, etc.) as our semiMVAE. For AMMSS, we tuned the parameters as suggested in~\cite{Cai2013Heterogeneous}. For AMGL and semiVAE, we used their default settings.
\subsection{Performance Evaluation}
To simulate semi-supervised learning scenario, on both datasets, we randomly labeled different proportions of samples in the training set, and remained the rest samples in the training set unlabeled.  For transductive semi-supervised learning, we trained models on the dataset consisting of the testing data and labeled data belonging to training set. For inductive semi-supervised learning, we trained models on the entire training set consisting of the labeled and unlabeled data. For supervised learning, we trained models on the labeled data belonging to training set, and test their performance on the testing set.
Table \ref{accuracy} presents the classification accuracies of all methods on SEED and DEAP datasets. The proportions of labeled samples in the training set vary from  $1\%$ to $3\%$.  Several observations can be drawn as follows.

\begin{table}[!ht]
\scriptsize
\caption{ Comparison with several supervised and semi-supervised methods on SEED and DEAP with few labels.  Results (mean$\pm$std) were averaged over 20 independent runs.}
\vskip -0.05in
\centering
\renewcommand{\arraystretch}{1.1}
\setlength\tabcolsep{5.0pt}
\label{accuracy}
\renewcommand{\multirowsetup}{\centering}
\begin{tabular}{|c|l|c|c|c|}
  \hline
  SEED data &Algorithms & 1\% labeled & 2\% labeled & 3\% labeled\\
  \hline
  \hline
  \multirow{3}{1.2cm}{Supervised learning}
  &MAE+SVM                       & .814$\pm$.031            & .896$\pm$.024           & .925$\pm$.024 \\
  &DCCA+SVM                       & .809$\pm$.035            & .891$\pm$.035           & .923$\pm$.028 \\
  &DCCAE+SVM                       & .819$\pm$.036            & .893$\pm$.034           & .923$\pm$.027 \\
  \hline
  \multirow{6}{1.2cm}{Transductive semi-supervised learning}
  &AMMSS                        & .731$\pm$.055            & .839$\pm$.036           & .912$\pm$.018  \\
  &AMGL                         & .711$\pm$.047            & .817$\pm$.023           & .886$\pm$.028 \\
  &MAE+TSVM                     & .818$\pm$.035            & .910$\pm$.025           & .931$\pm$.026 \\
  &DCCA+TSVM                    & .811$\pm$.031            & .903$\pm$.024           & .928$\pm$.021 \\
  &DCCAE+TSVM                   & .823$\pm$.040            & .907$\pm$.027           & .929$\pm$.023 \\
  &semiMVAE                    & \textbf{.861}$\pm$.037   & \textbf{.931}$\pm$.020  & \textbf{.960}$\pm$.021 \\
  \hline
  \multirow{4}{1.2cm}{Inductive semi-supervised learning}
  &semiVAE (Eye)                  & .753$\pm$.024            & .849$\pm$.055          & .899$\pm$.049 \\
  &semiVAE (EEG)                  & .768$\pm$.041            & .861$\pm$.040          & .919$\pm$.026 \\
  &semiVAE (Concat.)              & .803$\pm$.035            & .876$\pm$.043          & .926$\pm$.044  \\
  &semiMVAE                    & \textbf{.880}$\pm$.033   & \textbf{.955}$\pm$.020  & \textbf{.968}$\pm$.015 \\
  \hline
\end{tabular}
\vskip 0.08in
\renewcommand{\multirowsetup}{\centering}
\begin{tabular}{|c|l|c|c|c|}
  \hline
  DEAP data &Algorithms & 1\% labeled & 2\% labeled & 3\% labeled\\
  \hline
  \hline
  \multirow{3}{1.2cm}{Supervised learning}
  &MAE+SVM                       & .353$\pm$.027            & .387$\pm$.014           & .411$\pm$.016 \\
  &DCCA+SVM                      & .359$\pm$.016            & .400$\pm$.014           & .416$\pm$.018 \\
  &DCCAE+SVM                     & .361$\pm$.023            & .403$\pm$.017           & .419$\pm$.013 \\
  \hline
  \multirow{6}{1.2cm}{Transductive semi-supervised learning}
  &AMMSS                        & .303$\pm$.029            & .353$\pm$.024           & .386$\pm$.014  \\
  &AMGL                         & .291$\pm$.027            & .341$\pm$.021           & .367$\pm$.019 \\
  &MAE+TSVM                     & .376$\pm$.025            & .403$\pm$.031           & .417$\pm$.026 \\
  &DCCA+TSVM                    & .379$\pm$.021            & .408$\pm$.024           & .421$\pm$.017 \\
  &DCCAE+TSVM                   & .384$\pm$.022            & .412$\pm$.027           & .425$\pm$.021 \\
  &semiMVAE                    & \textbf{.424}$\pm$.020   & \textbf{.441}$\pm$.013  & \textbf{.456}$\pm$.013 \\
  \hline
  \multirow{4}{1.2cm}{Inductive semi-supervised learning}
  &semiVAE (Phy.)                 & .366$\pm$.024            & .389$\pm$.048          & .402$\pm$.034 \\
  &semiVAE (EEG)                  & .374$\pm$.019            & .397$\pm$.013          & .407$\pm$.016 \\
  &semiVAE (Concat.)              & .383$\pm$.019            & .404$\pm$.016          & .416$\pm$.012  \\
  &semiMVAE                    & \textbf{.421}$\pm$.019   & \textbf{.439}$\pm$.025  & \textbf{.451}$\pm$.022 \\
  \hline
\end{tabular}
\end{table}
First, the average accuracy of semiMVAE significantly surpasses the baselines in all cases.
Second, by examining semiMVAE against supervised learning approaches trained on very limited labeled data, we can find that semiMVAE always outperforms them. This encouraging result shows that semiMVAE can effectively  leverage the useful information from unlabeled data.
Third,  multi-view semi-supervised algorithms AMMSS and AMGL perform worst in all cases. We attribute this to the fact that graph-based shallow models AMMSS and AMGL can't extract the deep features from the original data.
Fourth, the performances of three TSVM based semi-supervised methods are moderate. Although MAE+TSVM, DCCA+TSVM and DCCAE+TSVM can also integrate multi-modality information from unlabeled samples, their two-stage learning can't obtain the global optimal model parameters.
Finally, compared with the single-view semi-supervised method semiVAE, our multi-view method is more effective in integrating multiple modalities.
%

\begin{figure}[!ht]
\vskip -0.07in
  \subfigure[SEED dataset] {\includegraphics[height=1.4in,width=1.6in]{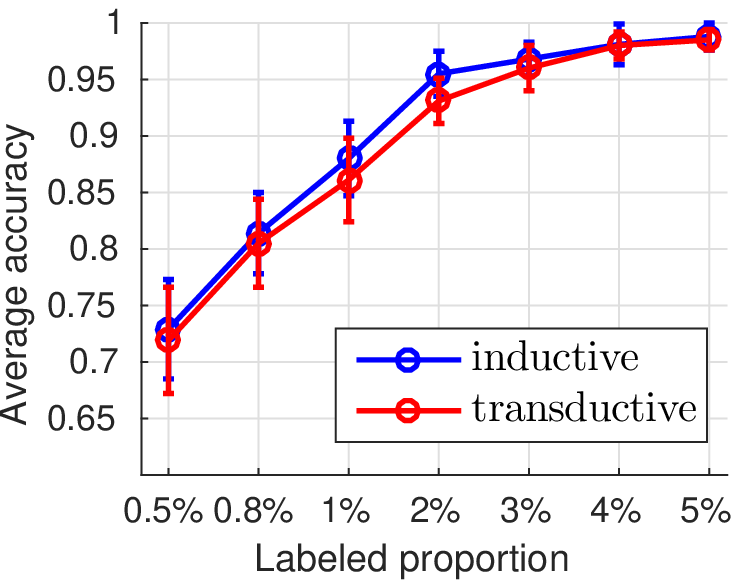}}
  \hspace {0.25cm}
  \subfigure [DEAP dataset] {\includegraphics[height=1.4in,width=1.6in]{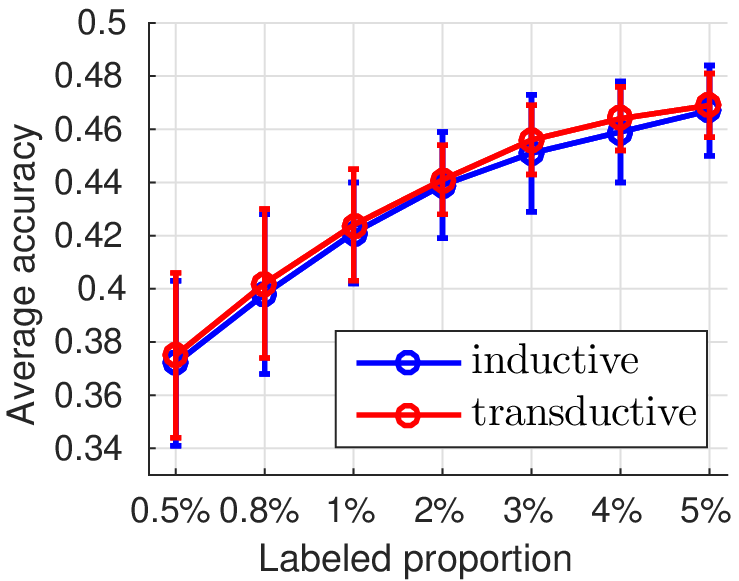}}
\vspace{-0.25cm}
\caption{semiMVAE's performance with different proportions of labeled samples in the training set.}
\label{fig:label_incre}
\end{figure}

\begin{figure}[!ht]
\vskip -0.01in
  \subfigure[SEED dataset] {\includegraphics[height=1.4in,width=1.6in]{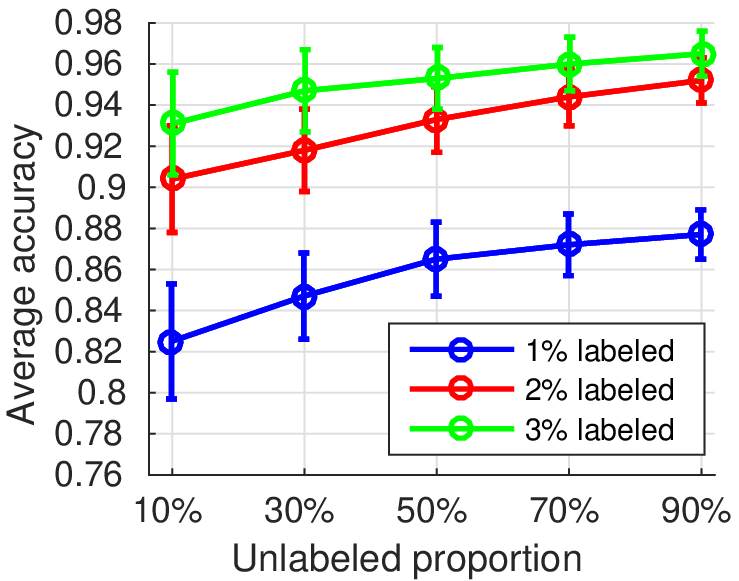}}
\hspace {0.25cm}
  \subfigure [DEAP dataset] {\includegraphics[height=1.4in,width=1.6in]{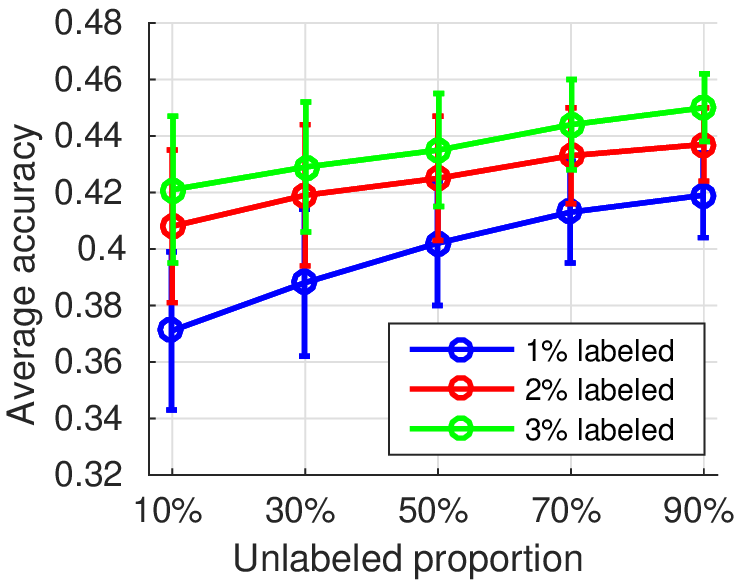}}
\vspace{-0.25cm}
\caption{Inductive semiMVAE's performance with different proportions of unlabeled samples in the training set.}
\label{fig:unlabel_incre}
\end{figure}
The proportion of labeled and unlabeled samples in the training set will affect the performance of semi-supervised models. Figs. \ref{fig:label_incre} and \ref{fig:unlabel_incre} show the changes of semiMVAE's average accuracy on both datasets with different proportions of labeled and unlabeled samples in the training set. We can observe that both labeled and unlabeled samples can effectively boost the classification accuracy of semiMVAE.

\begin{figure}[!ht]
\vskip -0.01in
  \subfigure[] {\includegraphics[height=1.4in,width=1.65in]{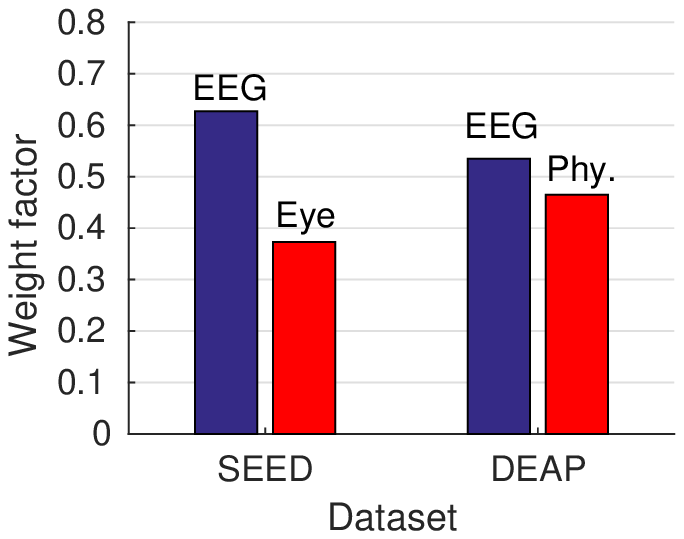}}
\hspace {0.20cm}
  \subfigure [] {\includegraphics[height=1.4in,width=1.59in]{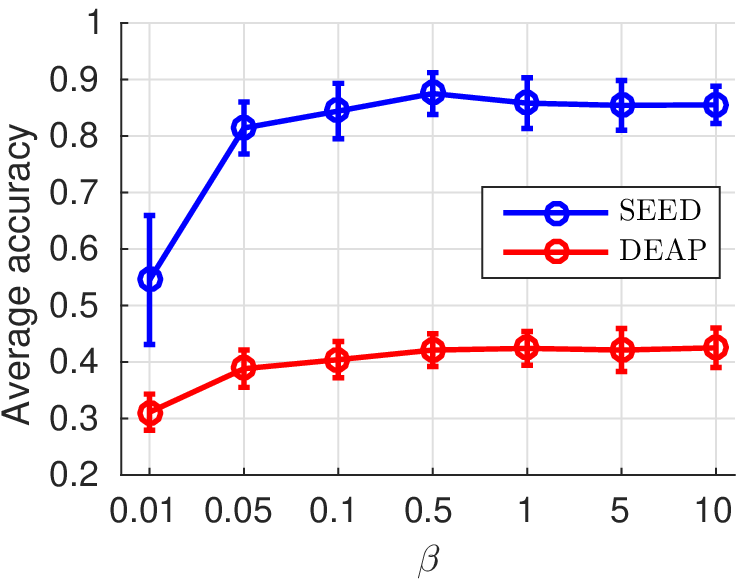}}
\vspace{-0.25cm}
\caption{(a) Learned weight factors by inductive semiMVAE.\ (b) The impact of scaling constant $\beta$.}
\label{fig:weight}
\end{figure}

Instead of treating each modality equally, our semiMVAE can weight each modality and perform classification simultaneously. Fig. \ref{fig:weight}a
shows the learned weight factors by inductive semiMVAE on SEED and DEAP datasets ($1\%$ labeled).  From it, we can observe that EEG modality has the highest weight on both datasets, which is consistent with single modality's performance of semiVAE shown in Table \ref{accuracy} and the results in previous work~\cite{lu2015combining}.

The scaling constant $\beta$ controls the weight of discriminative learning in semiMVAE. Fig. \ref{fig:weight}b shows the performance of inductive semiMVAE with different $\beta$ values ($1\%$ labeled). From it, we can find that the scaling constant $\beta$ can be chosen from \{0.1, 0.5, 1\}, where semiMVAE achieves good
results.
\section{Conclusion}
This paper proposes a semi-supervised multi-view deep generative framework for emotion recognition, which can leverage both labeled and unlabeled data from multiple modalities. The key to the framework are two parts: 1)  multi-view VAE can fully integrate the information from multiple modalities and 2) semi-supervised learning can overcome the labeled-data-scarcity problem. Experimental results on two real multi-modal emotion datasets demonstrate the effectiveness of our approach.

\smallskip\smallskip\smallskip
\vskip 0.3in

\newpage
\bibliographystyle{named}
\bibliography{changyingdu}

%
%

\end{document}